# Haul Road Mapping from GPS Traces

**Konstantin M. Seiler**

The University of Sydney

*k.seiler@acfr.usyd.edu.au*

Automation in mining requires accurate maps of road networks on site. Because roads on open-cut mines are dynamic in nature and continuously changing, manually updating road maps is tedious and error-prone. This paper investigates the possibility of automatically deriving an accurate representation of the road network using GPS data available from haul trucks operating on site. We present an overview of approaches proposed in literature and test the performance of publicly available methods on GPS data collected from trucks operating on site. Based on shortcomings seen in all tested algorithms, a post-processing step is developed which geometrically analyses the created road map for artefacts typical of free-drive areas on mine sites and significantly improves the quality of the final road network graph.

## 1. INTRODUCTION

Increasing automation in mining requires accurate maps of the roads on a mine site that accurately represent the topography and connectivity of the haul operation. Maps allow for detailed tracking of vehicle positions and states, as well as accurate planning of future movements by mine automation systems. Fleet management systems utilise road maps to compute routes and predict travel times between locations to maximise global objective functions such as total material moved, achieved blend, etc. The quality of the road map as a tool has a direct impact on truck queues and shovel hang times as well as product blend and crusher feeds.

Inferring a road network from positional data has been studied as map generation or map inference, and different approaches have been developed. However, these studies were conducted using massive datasets collected in urban environments, using consumer-grade GPS devices (Ahmed *et al.*, 2015; Biagioni and Eriksson, 2011). Thus, a big part of the challenge arises from working with large amounts of redundant but noisy data. This contrasts with a surface mine where fewer vehicles travel, GPS devices are typically of higher precision, and pathways change often as the operation progresses. Thus, while less data is available, it is generally more precise and accurate. Due to frequent changes in the physical road layout, it is not possible to compensate for the reduced amount of data by long observation periods. Map generation in this setting has not been studied in depth before. This study provides an overview of the available methods, evaluates their applicability to surface mining, and presents enhancements that overcome obvious deficiencies in the existing approaches.

Biagioni and Eriksson (2011) implement and quantitatively compare three map generation algorithms. Ahmed *et al.*, (2015) expand on this and review and benchmark a total of seven map generation algorithms, most of which have publicly available implementations. This work compares the performance of those algorithms and the newer method by Stanojevic *et al.*, (2018) on a dataset collected at an open-pit mine.

The remainder of this paper is structured as follows: Section 2 presents an overview of existing methods for road network inference from GPS data. Results of evaluating a selection of methods on data collected on a bauxite mine are presented in Section 3. Section 4 proposes a post-processing method to alleviate unwanted artefacts arising from situations typical in open-cut mining. Section 5 concludes.

**2. MAP INFERENCING APPROACHES**

Road map inferencing algorithms in literature broadly fall into one of four categories (Ahmed *et a*l., 2015; Biagioni and Eriksson, 2011). Direct incremental methods construct a map by iteratively adding the observed GPS traces to the road map graph while merging common road segments. Point clustering methods form spatial clusters out of the raw GPS points to create vertices of the road network, and infer the connectivity between the edges in a second step. Kernel Density Estimation (KDE) methods form a spatial histogram of the GPS points, often represented as a greyscale map, and then use methods related to computer vision to extract the road information. Intersection linking methods assume a grid-like urban road network and focus on identifying intersections which are subsequently linked by roads.

**2.1. Direct Incremental Methods**
Niehöfer *et al.,* (2009) build a road map by iteratively adding trips to the map. Duplicate road segments are identified by a proximity threshold and merged. The merging process forms a new path which is a weighted average of the location of the old path, resulting in increased accuracy of the road map as new data arrives. The merging process is such that the number of sample points along merged road segments increases, leading to a detailed description of the path, even when the individual sampling frequencies of the input traces are low.

Cao and Krumm (2009) generate a road map by iteratively adding individual trips. However, before building the final map, a clean-up step is performed. For the clean-up, a force-mass system is simulated. A variety of forces are employed to allow traces that belong to the same road to converge, while those belonging to different roads remain clearly separated. An experiment is conducted using GPS traces from 55 vehicles travelling across the Microsoft Campus in Seattle. The algorithm works well in low-error situations where the smoothing process is able to clarify the location of roads. In high-error situations, however, the smoothing process fails to match traces belonging to the same road, resulting in additional spurious roads (Biagioni and Eriksson, 2011).

Ahmed and Wenk (2012) present an incremental algorithm for map construction. While their algorithm doesn't attempt to update the position of edges and vertices to a common mean as additional data for a road segment arrives, edges are updated to reduce the complexity of the resulting graph. They provide proofs of the correctness of the resulting road map under a series of conditions, most notably a bound on the error found in GPS traces.

Tang *et a*l., (2017) present an incremental map construction algorithm which uses constraint Delaunay triangulation to refine the position of the road. Whenever a new trace is matched with an existing road in the road network graph, it is merged. A constraint Delaunay triangulation between the old path and the new is created and new vertices along the edges connecting the two paths are created according to the weights of the paths. This approach adds additional points to paths and thus allows for detailed recovery of the position and shape of the road.

**2.2. Point Clustering**
Edelkamp and Schrödl (2003) and Schrödl *et al.*, (2004) generate a road map by clustering GPS traces. Traces are selected as cluster seeds if they are far enough away in terms of location and heading from existing clusters. Close by traces are then clustered together. The individual traces are then linked using the connectivity information contained in the raw traces that pass through the cluster. The raw map that was obtained in this way is refined in a second step. For this, traces are matched to the created road segments and subsequently splines are fitted to best describe the centre line of the identified road. Lanes are separated by means of the heading information in the GPS traces. In case of noisy input traces, additional, spurious roads are added to the final road map. The algorithm has no means of detecting and pruning such false roads (Biagioni and Eriksson, 2011).

Stanojevic *et al.*, (2018) use a k-means clustering approach similar to Schrödl *et al.*, (2004). However, the spline fitting is omitted since k-means clustering already averages the positions of the individual traces. A sparsification step is performed to remove shortcuts that result from low sampling rates. Unlike previous k-means clustering approaches, Stanojevic *et al.*, (2018) also present an online version of the algorithm that allows for incrementally adding traces as they come in.

Ni *et al.*, (2018) combine a clustering approach with a refinement step based on Delaunay triangulations. Clusters are formed incrementally by adding each GPS point to the closest cluster seed and then updating the position of the cluster seed to the average of its contained points. For connectivity, the Delaunay triangulation of all cluster seeds is created and weights are assigned to the edges such that paths comprised of short edges are favoured. The shortest paths in the weighted Delaunay triangulation graph form the roads of the final road map.

### 2.3. Kernel Density Estimation

KDE based methods (Chen and Cheng, 2008; Davies *et al.*, 2006) discretise the workspace into cells. Individual GPS points are subsequently mapped to individual cells to form a spatial histogram. After application of a Gaussian smoothing function, the roads are recovered from the resulting image. This method works well when vast amounts of potentially noisy GPS data are available. Outliers are removed automatically and much-travelled roads show clearly. However, less travelled roads tend to lie below the threshold and are often not detected. Biagioni and Eriksson (2012) improve on this by creating a multitude of maps using different thresholds. The individual maps are then merged in a second step to produce a greyscale map where roads are colour-coded, based on their prominence in the data set. Post-processing interprets the greyscale map to prune spurious roads and derive the final road network graph.

### 2.4. Intersection Linking

Karagiorgou and Pfoser (2012) present a road map inference algorithm for urban road networks. The algorithm processes the input data in several stages to construct the road map. At first, positions where vehicles perform turns are identified in the GPS traces and clustered together. The clusters of turning points are assumed to be intersections and form the vertices of the road network. In a second step, the connectivity between intersections is deduced from the input GPS traces. A third step is performed for clean-up and merges redundant links between intersections to form the final output.

## 3. EXPERIMENTS

We tested some algorithms using publicly available implementations. For the experiments, truck movements of an Australian bauxite mine site were used. We chose this site because at the time of data collection the mine was small due to its young age, which offers advantages for this work: the entire road map can easily be shown in one figure containing main haul roads, benches, stockpile areas and the processing plant, making it easier to evaluate the map inference algorithms across different situations. While trucks were in motion, GPS positions were logged every six seconds. Positional data of a little over six days was used for construction of the road maps shown in this paper. The traces of each truck were split at gaps in the GPS data as well as when the speed dropped below 1 kph. All trips with more than 10 GPS points and 100 m of movement were used as input into the map construction algorithms. This resulted in 12807 trips with a total of 341025 individual GPS points.

### 3.1. Cao and Krumm (2009)

The algorithm of Cao and Krumm (2009) was re-implemented by Biagioni and Eriksson (2011) which is the version shown here. The algorithm is split into two separate parts. The first part performs several rounds of the 'clarification algorithm' which builds a spring-mass system where each GPS point is a mass object with a series of attracting and repelling forces between each of them. The positions of the waypoints are then updated in rounds according to the calculated forces reducing the effect of noise, and improving the resulting waypoint traces. The second part creates the road map by iteratively adding the clarified waypoint traces to the map.

Figure 1 (left) shows the resulting road map after only one round of clarification whereas Figure 1 (right) shows the map resulting from 39 clarification rounds. The algorithm recovers the overall road map reasonably well after only one round of clarification but there are some unwanted artefacts resulting from noisy GPS data or truck movements.

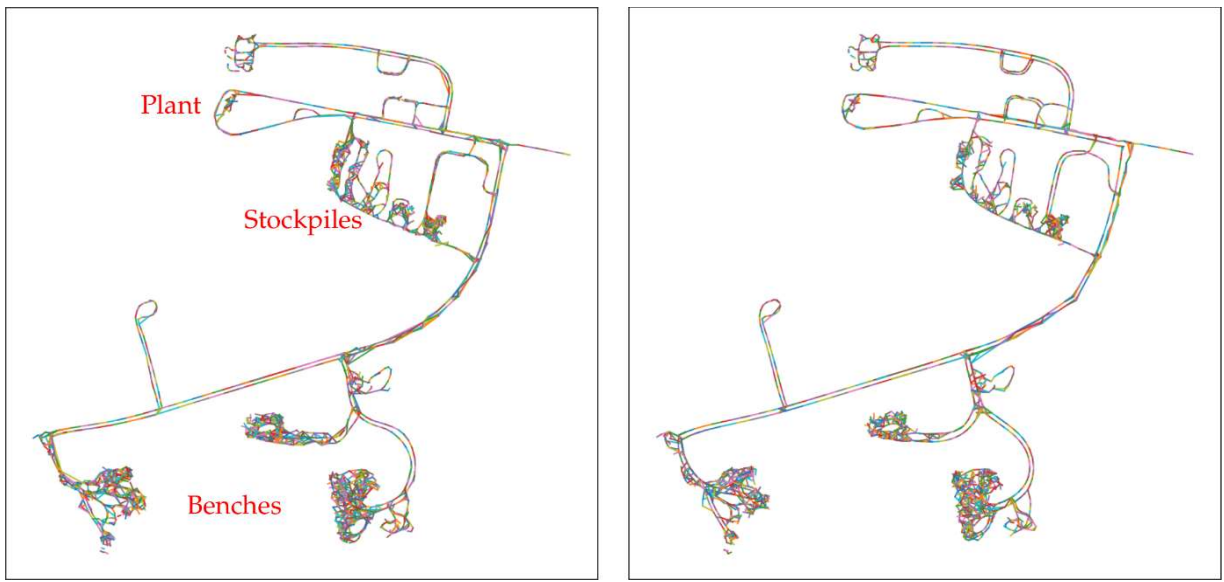

*Figure 1: Output of (Cao and Krumm, 2009) after one round of clarification (left) and 39 rounds of clarification (right). Each segment of the generated graph is coloured arbitrarily to better visualise the graph's structure.*

Most notable are the free-drive areas which show up as a messy conglomeration of unstructured roads. These are seen in the southern part of the map and arise predominantly on mining benches where well-defined roads are absent and trucks drive to their load location as needed. The stockpile area shows similar unwanted artefacts but to a lesser extent. These arise from vehicles not travelling on defined roads. They are expected, and appear in all methods that were tested here and manifest themselves in algorithmic dependent ways. Thus, the discussion will focus on the remainder of the map until an approach to clean up free-drive areas is presented in Section 4.

The algorithm fails to perform proper lane separation, resulting in spurious roads for lane changes being part of the graph. Intersection geometry is recovered poorly with some relevant connections missing, while others exhibit additional connecting road segments. Unfortunately, adding additional rounds of clarification doesn't seem to significantly improve the overall quality road map. On the contrary, the roads shift, and corners are cut, making the road map unsuitable for map matching. This is best visible when examining the loop road near the processing plant; its shape has changed significantly after the clarification rounds. Thus, the clarification stage appears to have little benefit for this dataset – due to the large scale of the roads and the high precision of the GPS units, measurement noise is less of an issue. Shifted road locations, while still topologically correct, quickly cause issues for applications that need to perform map matching.

### 3.2. Ahmed and Wenk (2012)

This algorithm is mostly an incremental method. Ahmed and Wenk (2012) describe a third stage of the algorithm which reduces the complexity of the resulting graph by constructing minimum link paths, but the code published by the authors shows no implementation of this step. It is unclear whether the review paper by the same authors (Ahmed *et al.*, 2015) considered the third stage at all. The results shown here are the output of the published code.

We ran the algorithm with different settings for $\epsilon$. Examples for $\epsilon = 15$ m and $\epsilon = 30$ m are shown in Figure 2. The algorithm captures the overall layout of the roads well. The road network that is constructed for the benches shows that the algorithm assumes that roads have a minimum distance of $\epsilon$ and the traces are sparsified accordingly. This is also the reason why the lanes are kept as distinct roads for $\epsilon = 15$ m and merged for $\epsilon = 30$ m. The road network shows some unexplained gaps along roads where the data should be sufficient. This is likely to be a bug in the implementation and not a shortcoming of the algorithm itself. The intersection geometry and connectivity isn't recovered properly, and artefacts regarding lane separation occur.

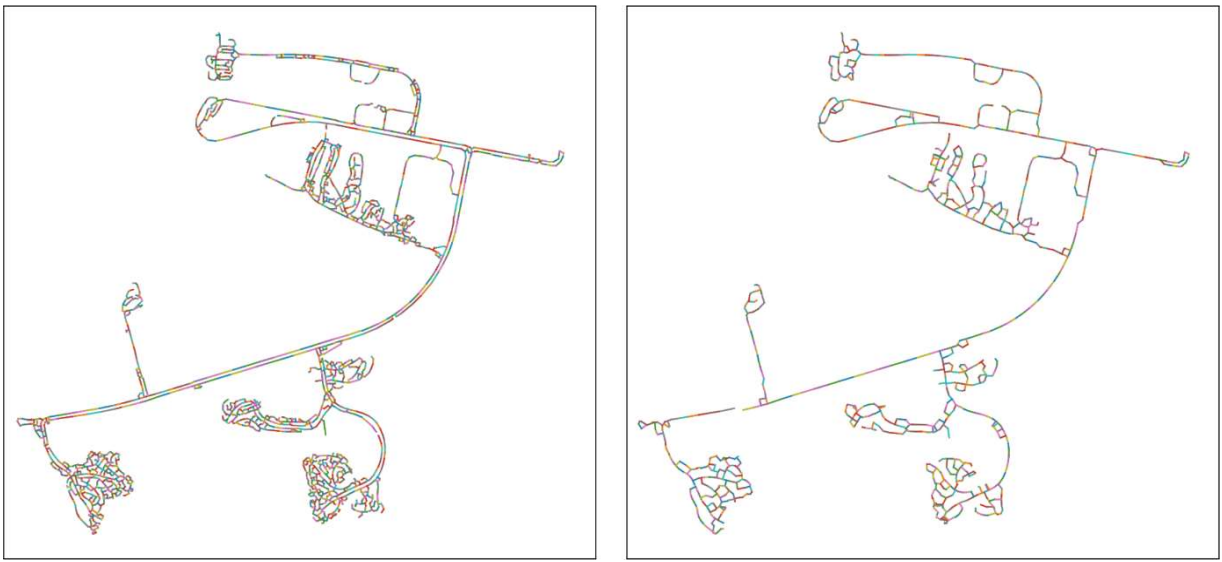

*Figure 2: Output of (Ahmed and Wenk, 2012) with $\epsilon = 15\ m$ (left) and $\epsilon = 30\ m$ (right).*

### 3.3. Davies et al. (2006)

This algorithm is a KDE based method. The implementation used here is a re-implementation by Biagioni and Eriksson (2011). The result is shown in Figure 3 (left). The method creates a greyscale image representing the density of GPS samples in any given location. From this the outline of the densely covered area is calculated (shown in black and red). In a final step the Voronoi diagram of the resulting polygons is created to form the final road graph (shown in blue).

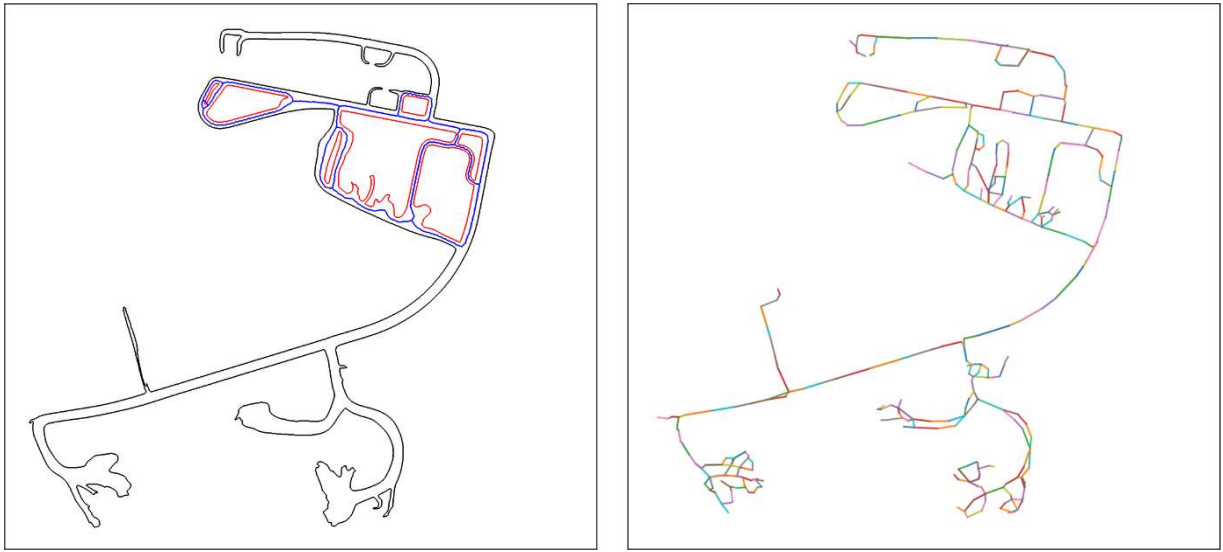

*Figure 3: Left: Output of (Davies et al., 2006). The actual road map is shown in blue, while road outlines are marked in black and red. Right: Output of (Biagioni and Eriksson, 2012).*

The output shows a series of failures on the tested dataset, the most prominent being that the final road graph only spans a tiny portion of the road map where the road graph is multiply connected. This is because the Voronoi diagram is only able to capture roads where the two sides are bounded by different boundary polygons. Thus, roads only bounded by the outer polygon (black) are missed.

The road boundary polygon itself misses roads, too. The map shows several loops or turning bays where the loop connection is missing. This is due to the used density threshold which only considers areas that have a high enough density of GPS points. This leads to a trade-off between filtering noise and omitting sparsely travelled roads.

The method fails to recover individual lanes and poorly recovers intersection geometries. When roads border free-drive areas as can be seen in the stockpile area, the inferred roads are pulled away from the main path and into the open area.

### 3.4. Biagioni and Eriksson (2012)
This KDE based algorithm attempts to overcome the threshold selection issue seen in (Davies *et al.*, 2006). It creates a multitude of maps with differing threshold settings and subsequently combines them with the final road map. The result is shown in Figure 3 (right). It performs substantially better than (Davies *et al.*, 2006) and the overall layout of the road network is recovered well. However, there are still roads missing, such as the incomplete loop to the left of the centre of the map and the broken loop in the top middle.

The unwanted artefact of pulling roads towards free-drive areas is no longer present. The method is unable to recover lanes and intersection geometries. The method produces considerably less noisy roads in free-drive areas than the other non-KDE methods. However, it is doubtful that the roads shown for free-drive areas are useful.

### 3.5. Edelkamp and Schrödl (2003)
The proposed in (Edelkamp and Schrödl, 2003; Schrödl *et al.*, 2004) aims to cluster similar GPS positions to vertices and then connect them using the connectivity information contained in GPS sequences. The implementation used here is the one published by the authors and the result for the mining dataset is shown in Figure 4 (left).

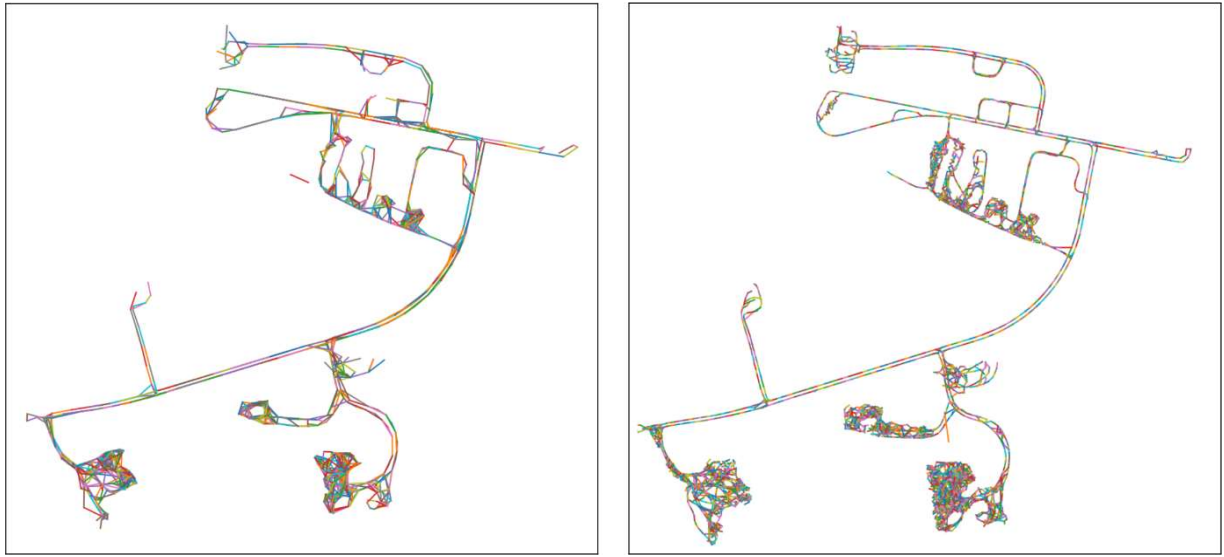

*Figure 4: Left: Output of (Edelkamp and Schrödl,2003). Right: Output of (Stanojevic et al., 2018) using a seed radius of 30 m.*

The algorithm spans most of the road network and separates lanes reasonably well, but it is noticeable that the algorithm fails to reliably detect duplicate roads. This can be best seen along bends on the main haul roads where multiple overlapping edges are visible in the road network graph. The same effect also leads to additional clutter around intersections.

The algorithm fails to detect some little-travelled roads as can be seen, for example at the incomplete loop top left of centre of the map.

### 3.6. Stanojevic et al. (2018)

The method by Stanojevic *et al*. (2018) was tested using the implementation published by the authors. The algorithm uses speed and heading of the GPS traces. Unfortunately, speed and heading information wasn't recorded for the data set used here for this experiment, and so it had to be calculated from consecutive GPS points. Potentially, the performance of the algorithm can be improved further by using natively recorded data. For this experiment, a seed radius of 30 m appears to be the most appropriate, and results from this setting are presented in Figure 4 (right).

The maps cover the travelled roads well, and lanes are well separated due to the use of heading information. The intersection geometry shown here is the best of the tested algorithms.

There are some artefacts in the form of cross-connected lanes, especially on the curved road segment leading to the bench at the bottom-centre of the map. The roads leading to the bench in the bottom-left of the map are not well defined.

Occasionally, duplicate road segments are present in the road map graph. However, this is hard to spot in the figure shown here. Duplicate roads arise when GPS traces aren't properly mapped onto the same sequence of clusters. If they aren't, nearby roads can appear. The algorithm contains a routine for detecting and pruning such paths, but unfortunately its design and/or implementation is flawed and it doesn't work as expected.

While the result is far from perfect, the good lane separation and intersection geometry produced by this algorithm are encouraging. Additional work towards cleaning up free-drive areas and cross-connected lanes is presented in Section 4. The method also implicitly cleans up duplicate roads where present. This, however, comes with the cost of spurious areas being marked.

### 3.7. Karagiorgou and Pfoser (2012)

The implementation of Karagiorgou and Pfoser (2012) was published by the authors and trialled on this dataset. Unfortunately, after several days of running time, the algorithm hadn't completed and was eventually aborted. It could be that the algorithm is very slow, or that some specifics of the dataset caused it to struggle. Since the algorithm was specifically geared towards urban networks with grid-like intersections, we did not further investigate the reason for failure.

### 4. MARKING AREAS

Given the results above, the method of Stanojevic *et al*. (2018) appears most promising. However, it still requires some further development to be able to generate maps suitable for use in automation systems. The most prominent issue is the arbitrary mess of roads within free-drive areas. The method presented here geometrically identifies free-drive areas, marks them as such and only keeps a small set of meaningful roads within them. The method also cleans up cross-lane connection artefacts and duplicate roads. It is designed to mark any intersections as areas, and subsequently simplify the connectivity of roads along intersections.

The algorithm stems from the observation that free-drive areas result in a local road network topology that contains many splits and merges of roads. Thus, every split or merge of a road is marked by a geometric shape in such a way that the shapes of splits and merges that belong to the same area overlap.

Once these areas are marked, the polygons and road network graph can be cleaned and pruned to derive a good representation of the road network. The basic shape used here for marking areas is a third of a filled circle which is anchored at the endpoint of an edge in the road network graph. An example of a graph overlaid with such shapes is shown in Figure 5. The radius of the circle section is 30 m and the opening angle is 120°. The shape is attached to i) the start of all outgoing edges that are not the only outgoing edge at a vertex, ii) the end of all incoming edges that are not the only incoming edge at a vertex and iii) the start or end of an edge that is the only edge at a vertex.

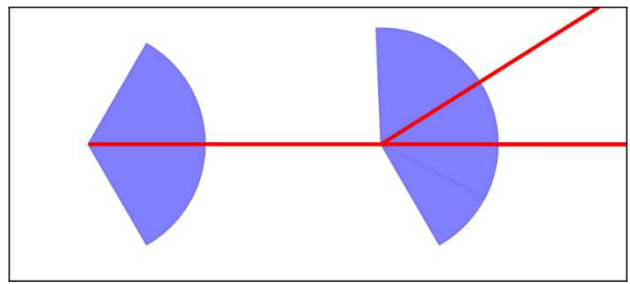

*Figure 5: Edges of the road map graph are marked with circle segments at end points and intersections.*

Once all marker shapes are placed, their geometric union is calculated to create a set of polygons. Circular arches are approximated by polygon shapes for computational reasons. Each polygon is then dilated by 11 m and subsequently eroded by 10 m. The effect of dilation with subsequent erosion is that narrow cuts and gaps in the polygons are filled to create a better representation of the area. The dilation distance is chosen slightly higher to ensure that intersection vertices of the road network are interior points of the multi-polygon. Note that the dilation-erosion process is done individually for each polygon so that polygons that overlap in their dilated state are not necessarily joined in the final result.

Following the initial calculation of areas, the road network is pruned, and areas are refined. This is performed in several rounds until no further changes are observed.

Vertices that lie within a marked area, and have an incoming edge coming from outside the area, are called entry nodes, and vertices that lie within a marked area, and have an outgoing edge that leaves the area are called exit nodes. First, all entry and exit nodes are identified and furthermore, the edges arriving at an area or leaving it are annotated with the area they come from or are loading to. Since all intersections are within areas, each such edge can only lead to exactly one other area. Then, the following steps are performed:

1. For each area, the entry and exit nodes of this area are considered. An attempt is made to find a path from each entry node to all exit nodes of the same area. If the exit node is directly next to the entry node in such a way that it constitutes the opposite lane of the same road, the exit node is skipped. The connecting paths are taken from the part of the road map graph which lies entirely within the area. In the likely case that multiple paths between an entry and exit node exist, the shortest path is chosen. Once all connections have been made, all internal edges of the marked area that don't belong to a shortest path are removed from the road map.

2. It is ensured that edges that connect two vertices from the same area are also part of the area, to avoid artefacts arising from concave boundaries. For this, all such edges that are not entirely within the area are identified geometrically. Then, the connecting edge is dilated by 5 m to turn it into a polygon shape and added to the area polygon by a geometric union.

3. All paths that lead from an area back into the same area without passing through another area are made part of the area by dilating the path by 5 m and adding it to the area polygons by a geometric union. This is similar to the previous step, but here there are intermediate vertices which lie outside of the area.

4. Areas that are closer together than 30 m and have more than one direct path connecting them in the same direction are connected by dilating the connecting paths by 5 m and adding them to the area polygons by a geometric union.

5. Polygons that overlap are merged by a geometric union.

6. Holes within resulting polygons are filled.

7. Each polygon is dilated and eroded by 10 m.

These steps are repeated until no more changes are being made in steps 2, 3 and 4. In practice the procedure terminates after only a few rounds.

This algorithm was used as a post-processing step for the result shown in Figure 4 (right). The resulting road map is shown in Figure 6.

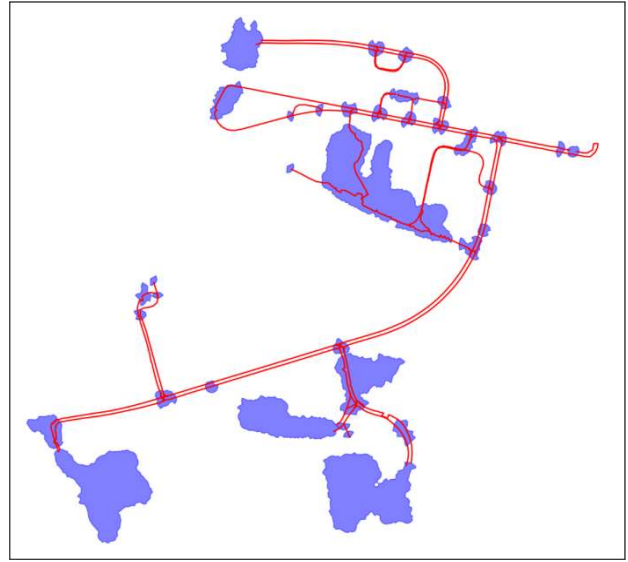

*Figure 6: Mining road network with marked areas.*

The road map shows the same overall coverage as before the post-processing step; however, the messy free-drive areas are replaced by marked areas, arguably a more sensible representation. In addition, intersections are also marked as areas. Cross-lane connections and duplicate roads are removed, but this comes at the expense of spurious areas being marked at those locations.

The method fails to create a meaningful representation of the stockpile area. Most of it is marked as a free-drive area, and the paths chosen within don't resemble the main road.

**5. CONCLUSION**

The tests show that none of the publicly available road network inference implementations are able to produce a road network of a quality that is adequate for use in automation systems in open-cut mines. Of the approaches tested, (Stanojevic *et al.*, 2018) seems to produce the best baseline road map due to its ability to separate lanes and recover intersection geometry.

It was shown that some of the shortcomings of existing methods, in particular (Stanojevic *et al.*, 2018), can be alleviated by targeted enhancements to the algorithm. This allows for improved road maps around turns and intersections, as well as marking open areas such as benches and stockpiles.

KDE-based approaches are less likely to perform well in a mine site setting because of the limited number of trips available on site and the high variation in travel frequencies across different parts of the road network. Furthermore, the strength of KDE approaches is their ability to deal with large amounts of noise within massive GPS datasets. However, noisy GPS data is much less of an issue on a mine site.

As shown in Figure 2, a purely incremental approach for adding trips to the graph while filtering for duplicate edges is promising. However, more work is required to remove the large amounts of unwanted artefacts that are still present in the generated graph.

The *k*-means-method by (Stanojevic *et al.*, 2018) is an in-between: clustering of the raw GPS points produces good location data and compensates for noise, whereas the consideration of heading and connectivity information present in individual trips allows for good graph generation with small datasets.

The algorithms discussed here are designed to batch-process a series of GPS traces in a single run. This is a valid approach when assuming a road network which is static over time. The challenge in a mining environment however is a dynamically changing road network. Using only data collected over a short period of time leads to less travelled roads being missed when creating the map, whereas larger collection periods lead to historic roads being indiscriminately overlapped with current ones. Thus, future work will investigate methods to continuously update a road map as the site progresses and new data becomes available.

**ACKNOWLEDGEMENT**

This work has been supported by the Australian Centre for Field Robotics and the Rio Tinto Centre for Mine Automation.